%% file: final_version.tex
\documentclass[acmtog]{acmart}

\usepackage{booktabs} 

\usepackage{multirow}
\usepackage{graphicx}

\usepackage[ruled]{algorithm2e} 
\usepackage{colortbl}  
\usepackage{hhline} 

\SetAlFnt{\small}
\SetAlCapFnt{\small}
\SetAlCapNameFnt{\small}
\SetAlCapHSkip{0pt}

\acmJournal{TOG}

\newcommand{\etal}{\textit{et al}.}

\newcommand{\xjqi}[1]{\textcolor[rgb]{1,0,0}{{[ #1]}}}
\newcommand{\xy}[1]{\textcolor[rgb]{0.1,0.8,0.5}{{{xy:} #1}}}

\definecolor{best}{rgb}{0.96, 0.57, 0.58}
\definecolor{second}{rgb}{0.98, 0.78, 0.57}
\definecolor{third}{rgb}{1.0, 1.0, 0.56}

\begin{document}

\setcopyright {acmlicensed} 
\acmJournal {TOG} 
\acmYear{2024} \acmVolume{43} \acmNumber{6} \acmArticle{198} \acmMonth {12} \acmDOI { 10.1145/3687952}

\title{3DGSR: Implicit Surface Reconstruction with 3D Gaussian Splatting}
\author{Xiaoyang Lyu}
\affiliation{
 \institution{The University of Hong Kong}
 \country{Hong Kong}
 }
\email{shawlyu@connect.hku.hk}
\author{Yang-Tian Sun}
\affiliation{
 \institution{The University of Hong Kong}
 \country{Hong Kong}}
\email{sunyt98@connect.hku.hk}
\author{Yi-Hua Huang}
\affiliation{
 \institution{The University of Hong Kong}
 \country{Hong Kong}}
 \email{huangyihua16@mails.ucas.ac.cn}
\author{Xiuzhe Wu}
\affiliation{
 \institution{The University of Hong Kong}
 \country{Hong Kong}}
 \email{xzwu@eee.hku.hk}
\author{Ziyi Yang}
\affiliation{
 \institution{The University of Hong Kong}
 \country{Hong Kong}
}
\email{14ziyiyang@gmail.com}
\author{Yilun Chen}
\affiliation{
 \institution{Shanghai AI Lab}
 \city{Shanghai}
 \country{China}
}
\email{chenyilun95@gmail.com}
\author{Jiangmiao Pang}
\affiliation{
 \institution{Shanghai AI Lab}
 \city{Shanghai}
 \country{China}
}
\email{pangjiangmiao@gmail.com}
\author{Xiaojuan Qi}
\authornote{Corresponding author}
\affiliation{
 \institution{The University of Hong Kong}
 \country{Hong Kong}
}
\email{xjqi@eee.hku.hk}

\begin{abstract}
In this paper, we present an implicit surface reconstruction method with 3D Gaussian Splatting (3DGS), namely 3DGSR, that allows for accurate 3D reconstruction with intricate details while inheriting the high efficiency and rendering quality of 3DGS. 
The key insight is to incorporate an implicit signed distance field (SDF) within 3D Gaussians for surface modeling, and to enable the alignment and joint optimization of both SDF and 3D Gaussians.
To achieve this, we design coupling strategies that align and associate the SDF with 3D Gaussians, allowing for unified optimization and enforcing surface constraints on the 3D Gaussians. With alignment, optimizing the 3D Gaussians provides supervisory signals for SDF learning, enabling the reconstruction of intricate details. However, this only offers sparse supervisory signals to the SDF at locations occupied by Gaussians, which is insufficient for learning a continuous SDF.
 Then, to address this limitation, we incorporate volumetric rendering and align the rendered geometric attributes (depth, normal) with that derived from 3DGS. 
In sum, these two designs allow SDF and 3DGS to be aligned, jointly optimized, and mutually boosted. 
Our extensive experimental results demonstrate that our 3DGSR  enables high-quality 3D surface reconstruction while preserving the efficiency and rendering quality of 3DGS.  Besides, our method competes favorably with leading surface reconstruction techniques while offering a more efficient learning process and much better rendering qualities.

\end{abstract}



\begin{teaserfigure}
  \centering
  \includegraphics[width=\textwidth]{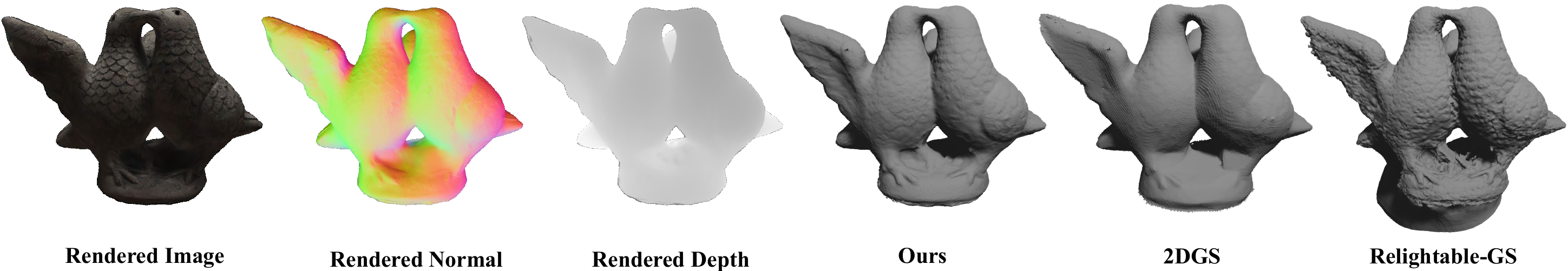}
  \caption{Our method, called 3DGSR, achieves accurate 3D surface reconstruction with rich details while maintaining the efficiency and high-quality rendering of 3DGS. The left part shows the capability of our method to achieve high-quality reconstruction and rendering results simultaneously. 2DGS \cite{Huang2DGS2024} is the state-of-the-art Gaussian-based reconstruction method.
  }
  \label{fig:teaser}
\end{teaserfigure}

\maketitle

\input{main/1.intro}
\input{main/2.related}
\input{main/3.preliminary}
\input{main/4.method}
\input{main/5.exp}
\input{main/6.conclusion}

\bibliographystyle{ACM-Reference-Format}
\bibliography{ref}

\input{./main/figure_only}

\end{document}

%% file: main/1.intro.tex
\section{Introduction}
3D Gaussian Splatting (3DGS) \cite{kerbl3Dgaussians} has emerged as a new state-of-the-art approach for high-quality novel view synthesis. This method represents the geometry and appearance of a 3D scene as a set of  Gaussians, which are then optimized from posed multi-view images.  
Thanks to its Gaussian rasterization pipeline, 3DGS achieves real-time efficiency, even when rendering high-resolution outputs.
Despite its impressive rendering quality and efficiency, 3DGS generates only noisy, incomplete point clouds for 3D geometry and struggles to accurately reconstruct the 3D surface of a scene~\cite{tang2023dreamgaussian}. These, however, are highly desired in various geometry-related applications, such as 3D reconstruction, geometry editing, animation, and relighting, among others. 

This motivates us to investigate how to enable 3DGS for high-quality surface reconstruction while preserving its rendering capabilities and speed. 
The challenges faced by 3DGS in accurately representing a 3D surface stem from (1) its unstructured point-based geometry representation (i.e., the center of Gaussian),  making it difficult to extract 3D surfaces through post-processing, such as Poisson surface reconstruction~\cite{kazhdan2013screened}; and (2) its elliptical shape, which is not naturally aligned with surface representations without proper constraints.
Furthermore, the optimization process does not consider the geometric constraints among the points that are indeed constrained by the underlying geometry (i.e., surfaces), leading to noisy points.  
The limited number of views utilized for learning further exacerbates this issue. 
\begin{figure}
    \centering
    \includegraphics[width=\linewidth]{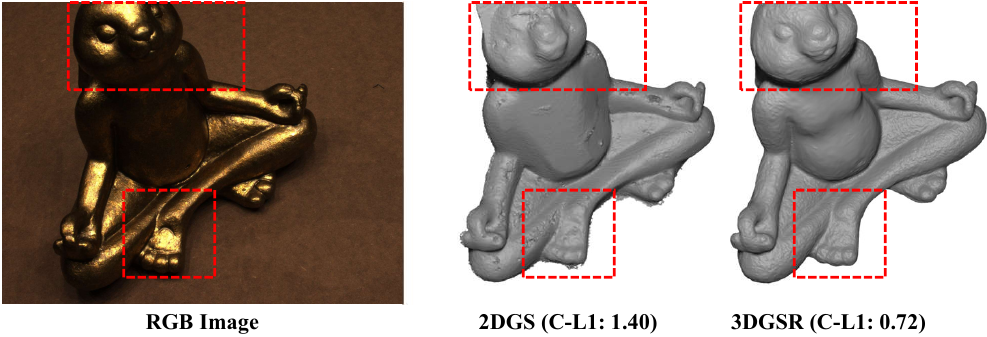}
    \caption{Oversmooth reconstruction results from 2DGS~\cite{Huang2DGS2024} lead to poorer performance, reflected by a higher chamfer-$\mathcal{L}_1$ error (1.40) compared to 3DGSR (0.72).}
    \label{fig: smooth}
\end{figure}

To this end, some recent works have explored the regularization of 3DGS~\cite{guedon2023sugar} or the use of 2DGS~\cite{Huang2DGS2024} or Gaussian surfels~\cite{Dai2024GaussianSurfels} for learning. Subsequently, a post-processing step, employing depth fusion~\cite{Huang2DGS2024} or Poisson reconstruction~\cite{Dai2024GaussianSurfels}, is required to obtain a 3D surface. Despite achieving promising results, these approaches tend to sacrifice detailed structures, resulting in smoother surfaces (Fig.~\ref{fig: smooth}) and suffering from limitations inherent to the reconstruction method.
Moreover, these formulations essentially decouple surface reconstruction from GS learning, which not only prevents the surface from regularizing the learning of GS but also leads to alignment issues and the accumulation of errors. 

\begin{figure}
    \includegraphics[width=0.95\linewidth]{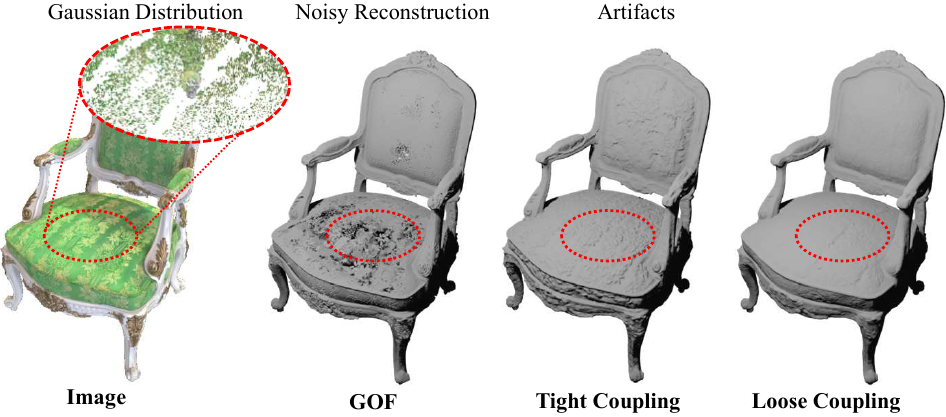}
    \caption{Imperfect reconstruction from the non-uniform distributions of Gaussians. Directly defining the surface from the Gaussian points may lead the noisy reconstruction or artifacts. }
    \label{fig: surface_def}
\end{figure}

In this paper,  we introduce 3DGSR, a simple yet effective method that integrates a lightweight neural implicit signed distance field (SDF) within Gaussians for surface modeling, which is optimized and jointly aligned with 3D Gaussians. 
We start by examining the alignment of an SDF field and 3DGS, enabling unified optimization and synchronization of the surface and appearance. 
Inspired by the intuition that Gaussians near the surface should possess a high opacity value, we explore a \textit{tight coupling} strategy that uses SDF to directly control Gaussian opacity with a differentiable SDF-to-opacity transformation function. 
Although this improves reconstruction quality, we find that surface reconstruction can be disrupted by high-frequency textures, particularly in low-frequency flat surface areas, leading to noisy reconstruction ({Fig.~\ref{fig: surface_def}}). This issue can also be observed in the concurrent work GOF~\cite{Yu2024GOF} ({Fig.~\ref{fig: surface_def}}), which directly leverages Gaussian opacity and volume rendering for surface extraction.  
Therefore, instead of tight coupling SDF and 3D Gaussians, we further investigate a \textit{loose coupling} strategy that only utilizes SDF to constrain the distribution and orientations of Gaussians. This approach forces the Gaussians to be near the surface derived from SDF and aligns their orientations (the direction of the shortest axis) with the normal of the surface. By doing so, we facilitate the alignment of surface and 3D Gaussians while providing flexibility for the SDF to model surface geometry and the 3D Gaussians to capture appearance details, yielding superior results in both rendering and reconstruction ({Fig.~\ref{fig:teaser}}).  

By employing the coupling strategies, we enable the optimization of Gaussians to supply supervisory signals for the SDF. However, these signals are only provided at locations occupied by the Gaussians, resulting in sparse supervision for the SDF. This sparsity proves insufficient for optimizing a continuous SDF and may lead to the creation of redundant structures.
To address this issue, we integrate volumetric rendering techniques~\cite{mildenhall2021nerf,wang2021neus} and align the rendered depth (or normal) with that obtained from the 3D Gaussians through a consistency loss. 
This approach effectively regularizes the regions not covered by the 3D Gaussians. 
It is worth noting that during each learning iteration, we only sample a few rays for rendering depth and normal using a single-layer MLP. This not only avoids introducing significant computational costs for training but also enables fast learning of SDF from the geometric cues provided by 3DGS. In turn, the improved SDF helps to regularize the 3D Gaussians effectively.

Extensive experiments conducted on multiple datasets demonstrate that our method can produce high-quality reconstruction results while maintaining the efficiency and rendering qualities of 3DGS. 
When compared to neural implicit reconstruction methods \cite{wang2021neus,yariv2021volume,tang2022nerf2mesh}, our approach performs on par or even surpasses them while being significantly faster and achieving superior novel view synthesis results. In comparison to GS-based methods \cite{Huang2DGS2024,Yu2024GOF}, our approach outperforms them in both surface reconstruction and novel view synthesis.

%% file: main/2.related.tex
\section{Related Work}

Our method builds upon prior works in 3D reconstruction with multi-view images to enable fast surface reconstruction and novel view synthesis. We discuss close related prior works in neural implicit representation and 3D Gaussian splatting (3DGS). 

\paragraph{Neural Implicit Representations}
In recent years, the fields implicitly predicted by neural networks have gained significant attention in both novel view synthesis~\cite{lombardi2019neural,sitzmann2019scene,mildenhall2021nerf} and 3D reconstruction~\cite{yariv2020multiview,niemeyer2020differentiable,kellnhofer2021neural,jiang2020sdfdiff,liu2020dist, yu2024gsdf3dgsmeetssdf}. Neural Radiance Field (NeRF)~\cite{mildenhall2021nerf} has demonstrated excellent performance in high-quality novel view synthesis by modeling scenes with a density field and a view-dependent color field. However, NeRF's density field cannot extract high-quality surfaces. To achieve accurate surface reconstruction, NeuS~\cite{wang2021neus} represents the scene with a signed distance function (SDF) and derives the alpha value from the precondition that the zero-level surface contributes the most to the volume rendering. Following NeuS, Darmon~\etal~\cite{darmon2022improving} improves the reconstruction quality with an additional warping loss that ensures photometric consistency. Neuralangelo~\cite{li2023neuralangelo} introduces multi-resolution hash grids~\cite{muller2022instant} and regularizes the model with numerical gradients and coarse-to-fine training. However, these methods often require extensive training time, which can limit their practical applicability.

\paragraph{3D Gaussian Splatting.}
Most recently,  3DGS~\cite{kerbl3Dgaussians} has demonstrated remarkable performance in terms of high-level rendering quality and real-time rendering. Subsequent works have extended 3DGS to handle 3D/4D generation~\cite{yi2023gaussiandreamer,tang2023dreamgaussian,zielonka2023drivable,ren2023dreamgaussian4d,yang2023learn}, dynamic scenes~\cite{luiten2023dynamic,yang2023deformable,yang2023gs4d, das2023neural, li2023spacetime,huang2023sc}, relighting~\cite{gao2023relightable}, and animation~\cite{jung2023deformable,ye2023animatable}.
Some concurrent works~\cite{guedon2023sugar, Huang2DGS2024, Yu2024GOF} aim to reconstruct the 3D surface from multi-view images using 3DGS. 
SuGaR~\cite{guedon2023sugar} enforces Gaussians to have limited overlap with others during the optimization of 3DGS. 
After that, an initial mesh can be extracted by performing Poisson reconstruction~\cite{kazhdan2013screened} on Gaussians. 
The mesh is then bound with flat Gaussians distributed on its faces and optimized to obtain refined results.
2DGS~\cite{Huang2DGS2024} represents the points as 2D surfels and leverages TSDF fusion to obtain the final mesh.
GOF~\cite{Yu2024GOF} leverages ray-tracing-based volume rendering of 3D Gaussians by identifying its level. Different from the SuGaR and 2DGS, they extract the mesh from a trained tetrahedral grid with marching tetrahedra.
Though these methods achieve promising reconstruction results, their surface still misses fine-grained geometry details and high-quality rendering results.
Here, we derive 3DGSR combined with explicit Gaussian rasterization and implicit volumetric rendering to get detailed surface reconstruction and high-quality rendering results.

%% file: main/3.preliminary.tex
\section{Preliminaries}\label{sec: pre}
\begin{figure*}
    \centering
    \includegraphics[width=\textwidth]{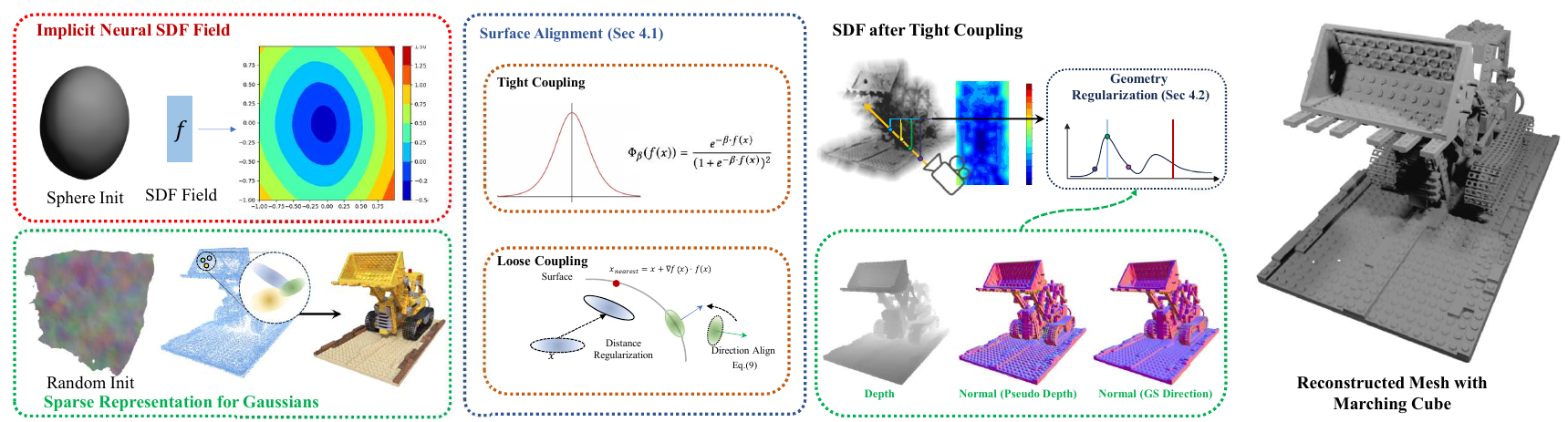}
\caption{Pipeline of our proposed approach for implicit surface reconstruction. We model the surface with an implicit SDF field, with which the SDF value of each 3D Gaussian can be predicted. We propose two different coupling strategies to make the distribution of Gaussians align with the implicit SDF field. The geometry attributes of 3D Gaussians serve as a regularization for the SDF field, while the rendered image is supervised by the captured image.}
\label{fig:pipeline}
\end{figure*}

\paragraph{3D Gaussian Splatting (3DGS)}
\label{sec:3dgs}
3DGS~\cite{kerbl3Dgaussians} employs 3D Gaussian points,  characterized by a central point $\mu$ and a covariance matrix $\Sigma$, to effectively render images from given viewpoints: 
\begin{equation}
    G(x) = \exp\left( -\frac{1}{2} (x - \mu)^T \Sigma^{-1} (x - \mu) \right),
\end{equation}
where $x$ is the point position in world space.

To facilitate optimization, $\Sigma$ is factorized into the product of a scaling matrix $S$, represented by scale factors $s$, and a rotation matrix $R$ encoded by a quaternion $r \in \mathbb{R}^4$:
\begin{equation}
\Sigma = RS S^T R^T,
\end{equation}
For rendering novel views, the 3D Gaussians are projected onto a 2D image plane according to elliptical weighted average (EWA)~\cite{zwicker2001surface}:
\begin{equation}
    \Sigma' = JW\Sigma W^T J^T,
\end{equation}
where $W$ is the viewing transformation matrix, and $J$ is the Jacobian derived from projective transformations. The final color $C(u)$ at pixel $u$ is rendered by combining $N$ ordered Gaussian splats using point-based $\alpha$-blending:
\begin{equation}
\label{eq:blending}
    C(u) = \sum_{i \in N} T_i \alpha_i c_i, \quad T_i = \prod_{j=1}^{i-1} (1 - \alpha_j).
\end{equation}
Here, $c_i$ is determined by the spherical harmonic (SH) function and the viewing direction, while $\alpha_i$ is computed from the 2D covariance $\Sigma'$, modulated by the opacity $\sigma$ of each Gaussian~\cite{kerbl3Dgaussians}.
In addition to rendering color $c_i$,  Eq.~\eqref{eq:blending} is also employed to render normal and depth as detailed below. 

\paragraph{Normal and Depth Estimation from Gaussian}
Typically, each 3D Gaussian can be treated as a point by using the centers of 3D Gaussians. Then, we can leverage the estimated depth and render them into a depth map using point-based alpha blending~\cite{gao2023relightable} as
\begin{equation}
    \mathcal{D} = \sum_{i \in N} T_i \alpha_i d_i,
    \label{eq: depth}
\end{equation}
where $d_i$ is the depth of $i$th point.
We then project the rendering depth into 3D space and get the pseudo normal from the local plane.

%% file: main/4.method.tex

\section{Method}
Given posed multiview images $\{\mathcal{I}_k\}$ and a point cloud obtained from the Structure from Motion algorithm~\cite{schoenberger2016sfm}, 
3DGSR integrates an implicit signed distance field (SDF) within 3D Gaussians (Fig.~\ref{fig:pipeline}) to reconstruct high-quality surfaces while preserving the rendering quality and efficiency of 3DGS. To achieve this, 3DGSR incorporates two key designs:
\textbf{Alignment of Surface and 3D Gaussians} (Sec.~\ref{subsec: alignment})  We investiage coupling strategies to synchronize surface and 3D Gaussians, enabling their alignment and joint optimization.  The \textit{tight coupling} strategy utilizes a differentiable SDF-to-opacity transformation function to connect the SDF field and the opacity of 3D Gaussians; and the \textit{loose coupling strategy} uses surface derived from SDF to constrain the distribution of Gaussians, make them near the surface and align the orientation of 3D Gaussians with surface normals.

By aligning 3D Gaussian and the surface, we allow unified optimization of 3D Gaussians and SDF. However, the supervisory signal only exists at locations occupied by 3D Gaussians and is thus not sufficient for supervising the dense SDF field. Thus, we propose the second design: \textbf{Regularization with Volumetric Rendering} (Sec. \ref{subsec: volume_constraint}). 
We employ volumetric rendering to render depth and normal by querying points along the cast ray from the neural implicit SDF and align them with that derived from 3D GS with a consistency regularization loss. 
This provides dense supervisory signals for learning the  SDF. The two designs work together to enforce surface constraints on 3D Gaussians, which in turn allows 3D Gaussians to assist in optimizing the SDF, resulting in high-quality surfaces with aligned 3D Gaussians. 

Compared to {previous} neural implicit surface reconstruction methods~\cite{wang2021neus, yariv2021volume}, our approach utilizes 3DGS for efficient and high-quality rendering while avoiding the dependence on time-consuming color prediction branches and volumetric rendering loss for optimization. Instead, it can be supervised by 3DGS and directly learn the SDF from optimized geometric clues (e.g., depth and normal) derived from 3DGS, making the training process more than 20 times faster (12 hours vs. 30 minutes).
In comparison to Gaussian-based reconstruction methods~\cite{Huang2DGS2024}, our approach unifies the optimization of 3DGS and an implicit surface, aligning them in an end-to-end manner. This eliminates the need for post-processing steps such as depth fusion or Poisson reconstruction and circumvents their limitations and error accumulations, leading to improved reconstruction quality and better alignment.

\subsection{Alignment of Surfaces and Gaussians}\label{subsec: alignment}

The vanilla 3DGS method optimizes millions of Gaussians to accurately fit the training views, enabling photorealistic synthesis of novel views. However, its main emphasis is on fitting the target views, which does not guarantee high-quality 3D geometries. 
Moreover, the 3D Gaussians should be constrained by the underlying geometry (i.e., surfaces) of the 3D scene, which is absent in existing methods. 

Here, we introduce neural implicit SDF, denoted as $f: \mathbb{R}^3 \rightarrow \mathbb{R}^1$ for surface modeling, which maps each 3D position $x\in \mathbb{R}^3$ to its signed distance from the surface. This is implemented using multi-resolution hash grids~\cite{muller2022instant} in conjunction with a single-layer MLP. 
The surface $\mathrm{S}$ of a scene can then be derived by using the zero-level set of its SDF, defined by: 
\begin{equation}
    \mathrm{S} = \{x \in \mathbb{R}^3 \: | \: f(x) = 0\}.
\end{equation}

Given the SDF $f$, the next question is to connect 3DGS with the implicit SDF to enable unified surface and appearance modeling and optimization. 
Here, we explore two coupling strategies: tight coupling with a differentiable SDF-to-opacity transformation function and loose coupling with distance and orientation regularization.

\paragraph{Tight Coupling Strategy.} 
Intuitively, points near the surface will have large opacity values and contribute more to rendering images. 
Thus, we introduce a differentiable SDF-to-opacity transformation function to tightly couple the SDF field and 3DGS. 
Consequently, we select a bell-shaped function $\Phi_{\beta}(f(x))$ controlled by a hyperparameter $\beta$, expressed as:

\begin{equation}
    \Phi_\beta(f(x)) = \frac{e^{-\beta \cdot f(x)}}{(1 + e^{-\beta \cdot f(x)})^2},
\end{equation}
where $\beta$ is a learnable parameter controlling the shape of the function.
This function transforms SDF values at a query location into the opacity of the corresponding Gaussian at that location. 
By linking the SDF with 3D Gaussians using a differentiable function, the SDF field can be updated through the optimization of 3D Gaussians by minimizing the photometric loss:
\begin{equation}
    \mathcal{L}_{c} = \lambda \mathcal{L}_\text{D-SSIM} + (1 - \lambda) \mathcal{L}_1,
\end{equation}
This loss function encourages the Gaussian at a location $x$ that contributes more to rendering a pixel to have a higher opacity value $\Phi_\beta(f(x))$. This, in turn, encourages the SDF $f(x)$ to be closer to $0$ and, therefore, nearer to the surface.

\paragraph{Loose Coupling.}
Although the tight coupling strategy can align Gaussian points with the implicit surface and enable unified optimization for enhanced reconstructions, it may slightly compromise rendering quality and produce unsatisfactory reconstruction results with visible artifacts (See Fig.~\ref{fig: surface_def}). Upon investigation, we find that the surface and color appearances indeed exhibit different levels of detail and frequencies, which could be a contributing factor.

For example, as shown in {Fig.~\ref{fig: surface_def}}, a chair with flat low-frequency surface areas may require numerous non-uniformly distributed Gaussian points to represent its high-frequency textures. In areas with dense Gaussian points, the tight coupling strategy enforces the SDF values to be closer to zero, potentially leading to noisy surface structures with unnecessary details (see Fig.~\ref{fig: ablation}) if the Gaussian locations are not accurate. This suggests that a tight alignment might not be the optimal solution.

Thus, we introduce a loose coupling strategy as an alternative to the tight coupling approach. The central concept involves using the surface derived from the SDF to regularize the distribution of 3D Gaussians. This strategy ensures that 3D Gaussians are positioned near the surface and their orientations are aligned with the surface normals. By aligning the surface normals instead of directly constraining the opacity, we can prevent interference with the supervision needed for high-frequency appearances.


Specifically, the direction of a 3D Gaussian at location $x_g$ is determined by the direction of its shortest axis of the covariance matrix, represented by $n_g$. The normal at $x_g$ in the SDF field is derived using $\nabla{f}(x_g)$. 
Note we normalize $\nabla{f}(x_g)$ to a unit vector.  
We then align the direction of the 3D Gaussian and the surface normal by minimizing the direction-aligned loss, which is defined as follows:
\begin{equation}
    \mathcal{L}_a = || 1 - n_g \nabla{f}(x_g) ||.
\end{equation}
Additionally, we constrain the 3D Gaussians to be near the surface. To achieve this, we query the SDF value of each Gaussian point $x_g$ and calculate its nearest surface location using the gradient of the SDF field as follows:
\begin{equation}
    x_{\text{nearest}} = x_g + f(x_g) \cdot \nabla{f}(x_g),
\end{equation}
where $x_{\text{nearest}} \in R$ represent the nearest 3D points of point $x_g$ on the surface from SDF. 
Then, we can use $\mathcal{L}_1 $ loss to minimize their distance. 
As shown in Fig.~\ref{fig: surface_def}, despite its simplicity, the loose coupling strategy yields detailed and accurate surface reconstruction results where the 3D Gaussians near the surface with their orientations aligned with surface normals. 
Moreover, this approach also helps reduce the number of Gaussians by preventing the generation of Gaussian points that are far away from the surface. Our experiments demonstrate that we can decrease the number of Gaussian points by 20\% without compromising, or even enhancing, the rendering quality.
If not explicitly specified, we use the loose coupling strategy for alignment by default. 





\subsection{Regularization with Volumetric Rendering}\label{subsec: volume_constraint}

By integrating the SDF field with 3D Gaussians and jointly training them, as discussed in Section~\ref{subsec: alignment}, we can achieve high-fidelity rendering results with an optimized SDF field. However, due to the sparse and discrete nature of 3D Gaussians, the supervision signals mentioned earlier are only available for locations occupied by 3D Gaussians, primarily near the surface. This limited availability is insufficient to regularize the continuous SDF, leading to artifacts in locations not covered by 3D Gaussians, such as floating artifacts in unoccupied free spaces, as shown in Fig.~\ref{fig:pipeline}. 

Volumetric rendering employs ray-casting to render a pixel,  sampling points along a ray that includes both occupied and free points. Thus, we propose incorporating volumetric rendering to optimize the dense SDF field. Specifically, we use volumetric rendering to render depth and normal, which can be directly derived from the SDF field, and align them with those derived from 3DGS for optimization.
There are two main reasons for using rendered depth and normal instead of color. First, rendering color would require an additional color prediction MLP network, significantly increasing the learning time. Second, and more importantly, learning the SDF directly from geometric clues (e.g., depth and normal) can alleviate the burden of appearance modeling, accelerate the learning process, and enhance the quality. 

More specifically, for each pixel, the ray cast from it is denoted as $\{ \mathbf{p}(t) = r_o + t\cdot r_d \: \| \: t>0\}$, where $r_o$ is the camera center and $r_v$ is the unit direction vector of the ray. 
We accumulate the depth $\hat{\mathbf{D}}(r)$ and normal $\hat{\mathbf{N}}(r)$ along the cast ray~\cite{wang2021neus}:
\begin{equation}
    \tilde{\mathbf{D}}(r) = \sum_{i=1}^{M}T^r_i\alpha_it^r_i \;, \tilde{\mathbf{N}}(r) = \sum_{i=1}^{M}T^r_i\alpha_i\tilde{n}^r_i \;,
\end{equation}
where $\tilde{n}$ is the normal computed by the gradient of SDF field $\triangledown f$, $T^r_i$ and $\alpha_i$ represent the transmittance and alpha value (a.k.a opacity) of the sample point, 
and their values can be computed by
\begin{equation}
        T^r_i = \prod_{j=1}^{i-1}(1-\alpha_i), ~
   \alpha_i = \max \left( \frac{\phi_s(f(x_i)) - \phi_s(f(x_{i+1}))}{\phi_s(f(x_i))}, 0 \right),\;
\end{equation}
where $\phi_s$ is the Sigmoid function.

After that, we utilize the depth $\mathcal{D}$ and normal $\mathcal{N}$ rendered by Gaussian Splatting to supervise the volumetric rendering results $\mathbf{D}$ and $\mathbf{N}$ by optimizing the following objectives:
\begin{equation}
    \mathcal{L}_\text{vd} = \sum_{\mathbf{r} \in \mathcal{R}}\|\mathcal{D}(r) - \tilde{\mathbf{D}}(r)\|_2 \;,
\end{equation}
and
\begin{equation}
    \mathcal{L}_\text{vn} = \sum_{\mathbf{r} \in \mathcal{R}}\|{\mathcal{N}}(\mathbf{r})-\tilde{\mathbf{N}}(\mathbf{r})\|_1+\|1-\mathcal{N}(\mathbf{r}) \tilde{\mathbf{N}}(\mathbf{r})\|_1\;,
\end{equation}
where $\mathcal{R}$ is the union of all training rays. 
These constraints regularize all the sampling points along the training rays, resulting in dense supervision for the implicit field and eliminating floating artifacts. 
As shown in Fig.~\ref{fig:pipeline}, there exist noisy SDF distributions (negative value) floating over the free space between the camera and the object, which is then removed by applying the regularization with volumetric rendering. Accurate geometry can be obtained through our careful training strategy and model design. 


\subsection{Training Objectives and Implementation Details}
To train 3DGRS, our objective is to get high-quality rendering and reconstruction results.
Thus, the final loss function is defined by
$$
\mathcal{L} = \mathcal{L}_c + \alpha_1\mathcal{L}_{vd} + \alpha_2\mathcal{L}_{vn} +\alpha_3\mathcal{L}_{a} + \alpha_4\mathcal{L}_{eik}
$$
where $\mathcal{L}_{eik}$ is the Eikonal equation~\cite{icml2020_2086} to penalize the deviation of the magnitude of grad $f$ from unit length:
\begin{equation}
    \mathcal{L}_\text{eik} = \mathbb{E}_x[(\| \triangledown f - 1 \|)^2].
    \label{eq: eikonal}
\end{equation}
In our experiments, we set $\alpha_1=1.0, \alpha_2=0.1, \alpha_3 = 0.0001, \alpha_4=0.1$.

We implement the 3DGSR with CUDA and tinycudann~\cite{tiny-cuda-nn}, building upon the framework of 3DGS. 
We customize the normal and depth maps CUDA rendering kernel.
During training, we increase the number of 3D Gaussian primitives following the adaptive control strategy in 3DGS.
During the volumetric rendering stage, we sample $N=1024$ rays per step.

For the synthetic data, we initialize the implicit function as a sphere, and the position of Gaussian is randomly initialized.
For the real-world scenes, which frequently exhibit intricate geometry and noisy poses, we adopt the SFM points as the initialized Gaussian position, and the network is still initialized as a sphere.
Furthermore, we also apply the sparse SFM points to ensure the consistency between the rendered depth $\mathcal{D}$ and projected depth map $D_{sfm}$ with $\mathcal{L}_2$ loss.
Specifically, when optimizing the pose using SFM algorithms like COLMAP, we can also obtain sparse 2D key points along with their corresponding 3D point locations. 
Given an image and its camera pose, we project \textit{visible} 3D points into the camera space and use the z-value as the depth $D_{sfm}$ for the corresponding 2D key points. We then apply an $\mathcal{L}_2$ loss to enforce consistency between the rendered depth $\mathcal{D}$ and the projected sparse depth $D_{sfm}$ based on the measurements.
Our models are trained for 30000 iterations, and all the experiments are conducted on a single NVIDIA RTX 3090 GPU.

%% file: main/5.exp.tex
\section{Experiments}
\subsection{Experimental Settings}
\paragraph{Datasets}
\begin{figure}
	\centering
	\includegraphics[width=0.95\linewidth]{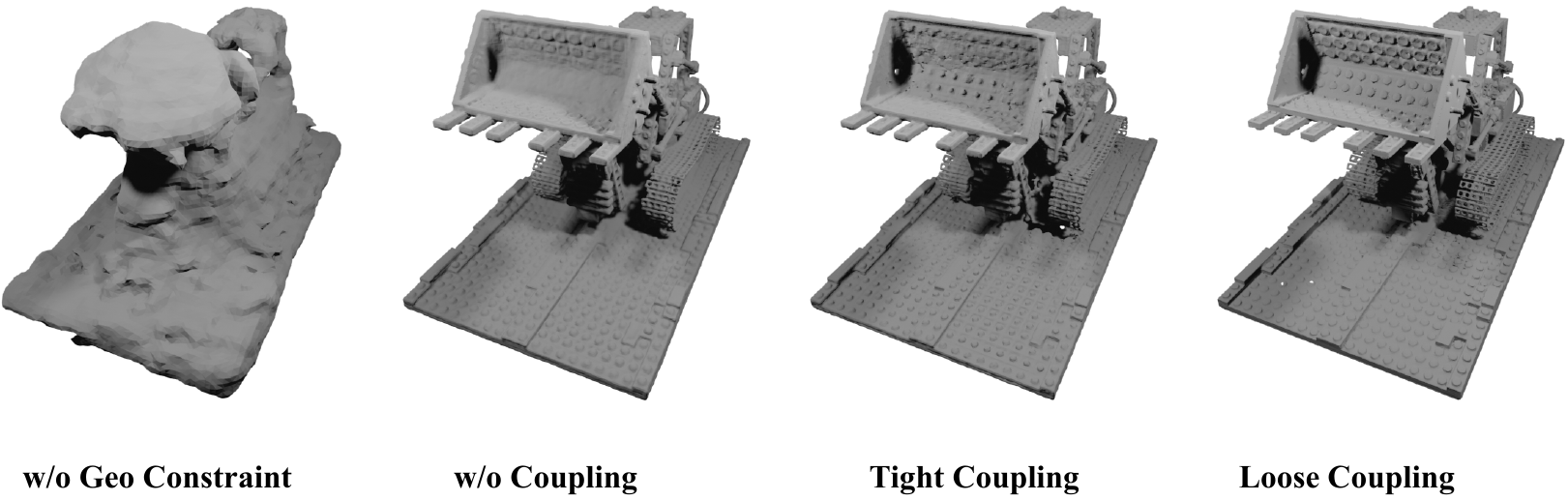}
	\caption{Qualitative comparisons of different coupling methods and the influence of geometric constraint.}
\label{fig: ablation}
\end{figure}

\begin{table*}[h]
  \centering
  \begin{tabular}{l|lccccccccc}
    \toprule[1pt]
                    & & Chair & Drums & Ficus & Hotdog & Lego & Materials & Mic & Ship & Avg\\
    \hline
    \multirow{9}{*}{\rotatebox[origin=c]{90}{PSNR}}  
    &NeRF~\cite{mildenhall2021nerf}              & 34.17 & 25.08 & 30.39 & 36.82 & 33.31 & 30.03 & 34.78 & 29.30 & 31.74 \\
    &Ins-NGP~\cite{muller2022instant}           & 35.00 & 26.02 & 33.51  & 37.40 & \cellcolor{best}36.39 &29.78 &36.22 &31.10 &33.18 \\
    &Mip-NeRF~\cite{barron2021mip}          & 35.14 & 25.48 & 33.29 & 37.48 & 35.70 & \cellcolor{best}30.71 & \cellcolor{second}36.51 & 30.41 & 33.09 \\
    &3D-GS~\cite{kerbl3Dgaussians}          & \cellcolor{second}35.36 & \cellcolor{second}26.15 & 34.87 & \cellcolor{second}37.72 & 35.78 & 30.00 & 35.36 
    & 30.80 & \cellcolor{second}33.32 \\
    &NeuS~\cite{wang2021neus}              & 31.22 & 24.85 & 27.38  & 36.04    & 34.06 & 29.59 & 31.56 & 26.94 & 30.20 \\
    &NeuS2~\cite{wang2023neus2}          &31.55 &24.65 &29.61 &34.84 &31.63  &27.68 &34.23  &28.92 & 30.39  \\
    &NeRO~\cite{liu2023nero}              & 28.74 & 24.88 & 28.38  & 32.13    & 25.66 & 24.85 & 28.64 & 26.55 & 27.48 \\
    &BakedSDF~\cite{yariv2023bakedsdf}       & 31.65 & 20.71 & 26.33  & 36.38    & 32.69 & 30.48 & 31.52 & 27.55 & 29.66 \\
    &NeRF2Mesh~\cite{tang2022nerf2mesh}         & 34.25 & 25.04 & 30.08  & 35.70    & 34.90 & 26.26 & 32.63 & 29.47 & 30.88\\
    & 2DGS~\cite{Huang2DGS2024}            & 35.05 & 26.05 & \cellcolor{second}35.57 & 37.36 & 35.10 & 29.74 & 35.09 & 30.60 & 33.07 \\
    &Ours (Tight Coupling) & 34.85 & 26.08 & 35.17 & 36.88 & 34.90 & 30.03 & 36.44 & \cellcolor{second}31.48 & 33.23 \\
    &Ours (Loose Coupling) & \cellcolor{best}35.69 & \cellcolor{best}26.35 & \cellcolor{best}35.65 & \cellcolor{best}37.98 & \cellcolor{second}36.19 & \cellcolor{second}30.31 & \cellcolor{best}37.12 & \cellcolor{best}31.65 & \cellcolor{best}33.86 \\
    \bottomrule[1pt]
    \multirow{7}{*}{\rotatebox[origin=c]{90}{C-$\mathcal{L}_1$}} 
    &VolSDF~\cite{yariv2021volume}              & 1.18 & 4.03 & 3.01  & 3.22    & 2.26 & 1.57 & 1.13 & 6.42 & 2.86 \\   
    &NeuS~\cite{wang2021neus} & 3.99 & 1.27 & 0.94 & 2.12 & 2.56 & 1.39 & 1.00 & 5.38 & 2.33 \\
     &NeRO~\cite{liu2023nero}              & 1.27 & 1.97 & 1.22  & \cellcolor{second}1.88    & 1.90 & \cellcolor{best}1.33 & 0.87 & 4.95 & 1.92 \\
    &BakedSDF~\cite{yariv2023bakedsdf}              & 1.83 & 1.43 & 1.09 & \cellcolor{best}1.68    & \cellcolor{best}1.13 & 1.36 & \cellcolor{second}0.84 & 3.88 & 1.66 \\
    &NeRF2Mesh~\cite{tang2022nerf2mesh} & 1.62 & 1.11 & \cellcolor{best}0.65 & 2.73 & 1.93 & 1.42 & \cellcolor{best}0.78 & \cellcolor{best}2.20 & 1.55 \\
    &RelightableGaussian~\cite{gao2023relightable} & 3.65 & 2.34 & 1.26 & 3.11 & 1.63 & \cellcolor{second}1.35 & 1.76 & 3.35 & 2.31\\
    &Ours (Tight Coupling) & \cellcolor{second}1.01 & \cellcolor{second}0.95 & 0.69 & 1.92 & 1.35 & \cellcolor{second}1.35 & 1.15 & 3.35 & \cellcolor{second}1.50 \\
    &Ours (Loose Coupling) & \cellcolor{best}0.99 & \cellcolor{best}0.93 & \cellcolor{second}0.68 & 1.90 & \cellcolor{second}1.32 & \cellcolor{best}1.33 & 1.15 & \cellcolor{second}2.64 & \cellcolor{best}1.37 \\
    \bottomrule[1pt]
  \end{tabular}
  \caption{We assess the quality of synthesized images and the accuracy of surface reconstruction, with each cell colored to indicate the \colorbox{best}{best} and \colorbox{second}{second best}. Our method is compared against various state-of-the-art (SOTA) approaches in tasks of novel view synthesis and surface reconstruction. It outperforms all competitors in both tasks, achieving the highest PSNR and the lowest Chamfer-$\mathcal{L}_1$ (C-$\mathcal{L}_1$) distance.}
  \label{tab: all in NeRF Synthetic}
\end{table*}
Our evaluation encompasses three datasets: the NeRF synthetic dataset~\cite{mildenhall2021nerf}, the DTU real-captured dataset~\cite{jensen2014large}, and the Tanks \& Temples (TNT) large-scale dataset~\cite{knapitsch2017tanks}.
The NeRF synthetic dataset consists of eight objects, each rendered via path tracing to generate 100 views.
The DTU and TNT datasets are processed to obtain camera poses in accordance with~\cite{gao2023relightable}, while the images and masks are processed following the methodology outlined in~\cite{liu2020neural}. The image resolutions for the DTU and TNT datasets are $800 \times 600$ and $1920 \times 1080$, respectively. 

\paragraph{Comparisons}
We benchmark our method on two tasks: surface reconstruction and novel view synthesis. Our approach is compared against several state-of-the-art \textbf{1) Neural surface reconstruction} methods, including NeuS~\cite{wang2021neus}, NeuSG~\cite{chen2023neusg}, RelightableGaussian~\cite{gao2023relightable}, NeRF2Mesh~\cite{tang2022nerf2mesh}, Neuralangelo~\cite{li2023neuralangelo}, NeUS2~\cite{wang2023neus2}, BakedSDF~\cite{yariv2023bakedsdf}, VolSDF~\cite{yariv2021volume}, and \textbf{2) Gaussian-based reconstruction methods} such as SuGaR~\cite{guedon2023sugar}, 2DGS~\cite{Huang2DGS2024}, and GOF~\cite{Yu2024GOF}. It is worth noting that GOF~\cite{Yu2024GOF} is a concurrent work.
Additionally, we also compare the novel view synthesis capabilities of our method with works that focus solely on \textbf{novel view synthesis} tasks, including NeRF~\cite{mildenhall2021nerf}, Mip-NeRF~\cite{barron2021mip}, Ins-NGP~\cite{muller2022instant}, NeUS2~\cite{wang2023neus2}, and 3DGS~\cite{kerbl3Dgaussians}. The quality of image synthesis is quantified using PSNR. For surface reconstruction accuracy, we employ Chamfer distance and the F1 score~\cite{knapitsch2017tanks}.

\begin{table*}[h]
\centering
\small
\begin{tabular}{@{}lccccccccccccccccc@{}}
\toprule
{Scan ID}     & {24} & {37} & {40} & {55} & {63} & {65} & {69} & {83} & {97} & {105} & {106} & {110} & {114} & {118} & {122} & {Mean} \\ \hline
 NeRF~\cite{mildenhall2021nerf} & 1.90 & {1.60} & {1.85} & {0.58} & {2.28} & {1.27} & {1.47} & {1.67} & {2.05} & {1.07} & {0.88} & {2.53} & {1.06} & {1.15} & {0.96} & {1.49}\\
NeuS \cite{wang2021neus}    & 1.00        & 1.37        & 0.93        & 0.43        & 1.10        & \cellcolor{second}0.65        & \cellcolor{second}0.57 & 1.48        & \cellcolor{third}1.09        & 0.83         & \cellcolor{second}0.52         & 1.20         & \cellcolor{second}0.35         & 0.49         & 0.54         & 0.84       \\

VolSDF \cite{yariv2021volume}  & 1.14        & 1.26        & 0.81        & 0.49        & 1.25        & \cellcolor{third} 0.70        & 0.72        & 1.29        & 1.18        & \cellcolor{third}0.70         & 0.66         & 1.08         & 0.42         & 0.61         & 0.55         & 0.86      \\    
NeuS2 \cite{wang2023neus2}    & 0.56       & \cellcolor{second}0.76        & 0.49        & \cellcolor{third}0.37        & \cellcolor{second}0.92        & 0.71        & 0.76 & 1.22        & \cellcolor{second}1.08        & \cellcolor{best}0.63         & 0.59         & \cellcolor{third}0.89         & 0.40         & \cellcolor{second}0.48         & 0.55         & \cellcolor{second}0.70         \\
Neuralangelo \cite{li2023neuralangelo}  &  \cellcolor{best}0.37        & \cellcolor{best}0.72        & \cellcolor{best} 0.35        & \cellcolor{best} 0.35       &  \cellcolor{best} 0.87        &  \cellcolor{best}0.54       & \cellcolor{best}0.53        & 1.29        & \cellcolor{best}0.97        &   0.73         & \cellcolor{best}0.47         &\cellcolor{second}0.74         & \cellcolor{best}0.32         & \cellcolor{best}0.41         & \cellcolor{second}0.43         & \cellcolor{best}0.61      \\ \hline
ReGS \cite{gao2023relightable}  & 1.21        & 1.00        & 1.10        & 0.87        & 1.03        & 1.51        & 1.51        & \cellcolor{best}1.06        & 1.63        & 0.97         & 1.36        & 1.21         & 0.94         & 1.38         & 1.26         & 1.20    \\ 
SuGaR \cite{guedon2023sugar}  & 1.47        & 1.33        & 1.13        & 0.61        & 2.25        & 1.71        & 1.15        & 1.63        & 1.62        & 1.07         & 0.79        & 2.45         & 0.98         & 0.88         & 0.79         & 1.33        \\
2DGS \cite{Huang2DGS2024}  & 0.48        & 0.91        & \cellcolor{third}0.39        & 0.39        & \cellcolor{third} 1.01        & 0.83        & 0.81        & 1.36        & 1.27        & 0.76         & 0.70        & 1.40         & 0.40         & 0.76         & 0.52         & 0.80       \\
GOF \cite{Yu2024GOF} (concurrent)  & \cellcolor{third}0.50        &  \cellcolor{third}0.82        & \cellcolor{second}0.37        & \cellcolor{third}0.37        & 1.12        &  0.74        & 0.73        & \cellcolor{second}1.18        & 1.29        & \cellcolor{second}0.68         & 0.77        & 0.90         & 0.42         & 0.66         & \cellcolor{third}0.49         & \cellcolor{third}0.74        \\
Ours & \cellcolor{second}0.44 & 0.96 & 0.40 & \cellcolor{second}0.36 & 1.02 & 0.80 & \cellcolor{third}0.64 & \cellcolor{third}1.20 & \cellcolor{second}1.08 & 0.97 & \cellcolor{third}0.54 & \cellcolor{best}0.72 & \cellcolor{third}0.37 & \cellcolor{third}0.52 & \cellcolor{best}0.42 & \cellcolor{second}0.70 \\
\bottomrule
\end{tabular}
\caption{Quantitative assessment on the DTU dataset with each cell colored to indicate the \colorbox{best}{best}  \colorbox{second}{second best} and \colorbox{third}{third best}. Our method achieves the highest average quality of surface reconstruction and the lowest Chamfer-$\mathcal{L}_1$ distance with real-world data.}
\label{tab: all in dtu}
\end{table*}
\begin{figure}
	\centering
	\includegraphics[width=1.\linewidth]{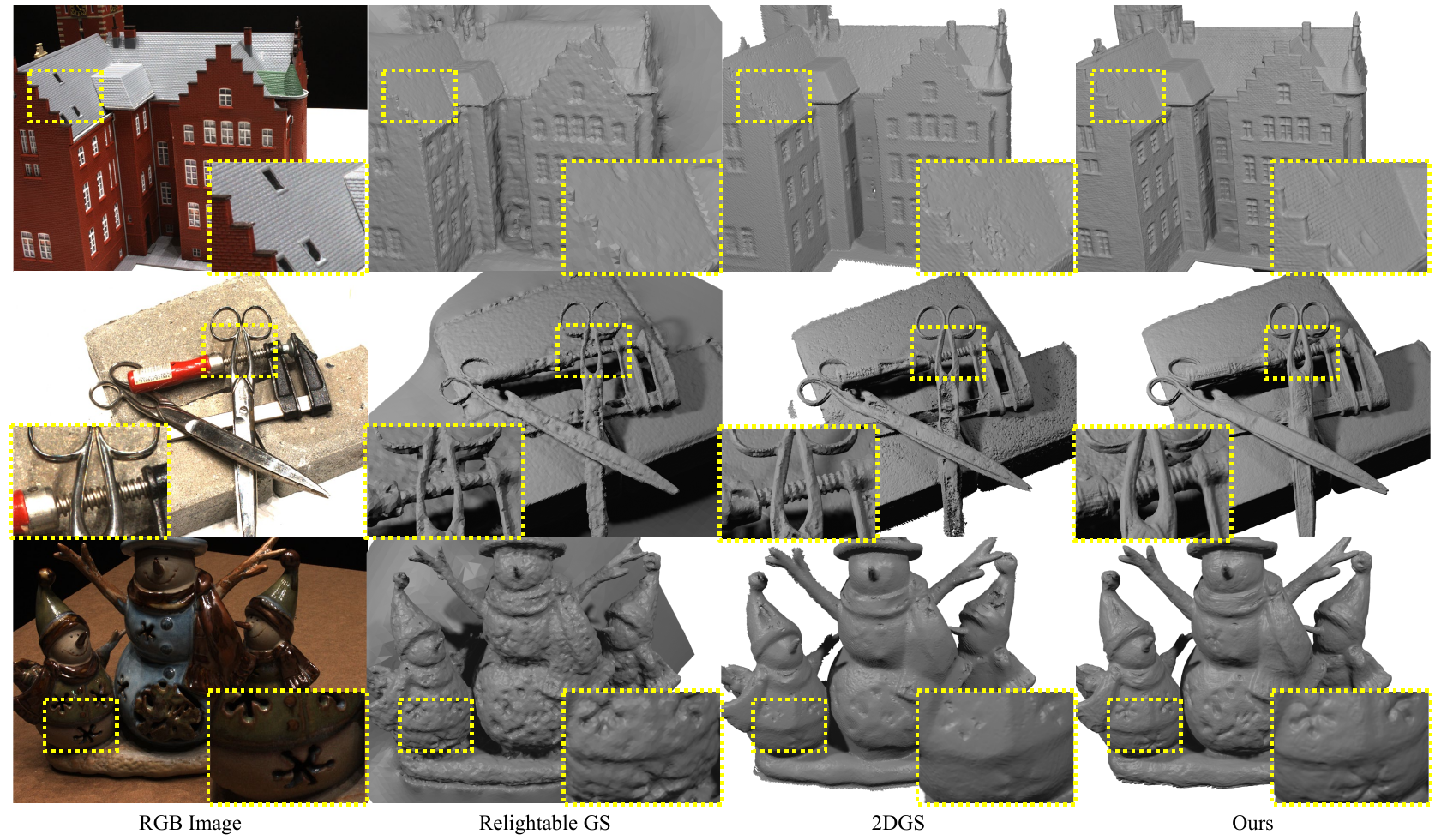}
	\caption{Qualitative comparisons of reconstructed results between Relightable 3D Gaussian~\cite{gao2023relightable}, 2DGS~\cite{Huang2DGS2024} and our method on the DTU dataset. Our method yields the highest-quality surface reconstruction preserving fine details and intricate structures.}
    \label{fig: DTU}
\end{figure}
\begin{table}[h]
    \centering 
    \begin{tabular}{lccc}
        \toprule
        Method & PSNR $\uparrow$ & SSIM $\uparrow$& LPIPS $\downarrow$\\ 
        \hline
        NeRF~\cite{barron2023zipnerf} & 31.74 & 0.947 & 0.056\\
        NeUS2~\cite{wang2023neus2} & 30.39 & 0.944 & 0.064\\
        3D-GS~\cite{kerbl3Dgaussians} & 33.32 & \textbf{0.970} & \textbf{0.031} \\
        2DGS~\cite{Huang2DGS2024}  & 33.07 & 0.967 & 0.033\\
        Ours  & \textbf{33.86}   & \textbf{0.970}  & \textbf{0.031} \\
        \bottomrule
    \end{tabular}
    \caption{More quantitative assessment on the NeRF synthetic dataset.}
    \label{tab: lpips}
\end{table}

\subsection{Main Results and Comparisons} 
\paragraph{Quantitative analysis} Tables~\ref{tab: all in NeRF Synthetic},~\ref{tab: all in dtu} and~\ref{tab: lpips} present the quantitative results for NeRF-synthetic and DTU datasets, respectively. 
On the NeRF-synthetic dataset (Table~\ref{tab: all in NeRF Synthetic} and Table~\ref{tab: lpips}), our approach achieves the best performance in both 3D surface reconstruction and novel view synthesis. Notably, our approach even outperforms the leading neural surface reconstruction methods for reconstruction evaluation while being much faster in training and yielding much better novel view synthesis results. 
Besides, our approach also delivers higher novel-view synthesis results than 3DGS \cite{kerbl3Dgaussians} and 2DGS \cite{Huang2DGS2024} in terms of PSNR, while achieving performance comparable to 3DGS on SSIM and LPIPS.
On the DTU dataset (Table~\ref{tab: all in dtu}), our method also exhibits the best performance in terms of average surface reconstruction quality when compared with Gaussian-based reconstruction methods. 
It also performs favorably when compared with leading neural implicit reconstruction methods, which take much longer time to train and often yield inferior rendering qualities.
We find that existing methods that achieve better reconstruction qualities tend to sacrifice the rendering qualities with lower performance than the NeRF baseline. 
This might be attributed to the different levels of detail required by surface reconstruction and novel view synthesis. 
Overall, our 3DGSR achieves excellent performance in both surface reconstruction and novel view synthesis.  
\paragraph{Qualitative analysis} 
We provide qualitative comparisons of the reconstruction results in Figures~\ref{fig: DTU} and~\ref{fig:our_mesh_comparison}. The state-of-the-art NeRF-based 3D reconstruction method, NeuS \cite{wang2021neus}, fails to preserve fine details such as intricate Mic textures, Lego tracks, and stripes on the Hotdog disk. The 2DGS method \cite{Huang2DGS2024}, which utilizes 2D Gaussians and relies on an additional post-processing depth fusion step to obtain 3D surfaces, tends to produce over-smoothed reconstruction results, as evidenced by the smoothed Lego track and house.
GOF \cite{Yu2024GOF} employs Gaussian opacities and volumetric rendering to extract meshes directly from 3DGS, unifying geometry and appearance modeling for 3DGS. However, due to the inherent differences in the level of detail for geometry and appearance modeling, their results tend to produce noisy and non-existent structures needed for modeling high-frequency appearances, as seen in the noisy balls and Lego tracks in Figure \ref{fig:our_mesh_comparison}.
In contrast, our approach is capable of producing high-quality reconstructions, effectively preserving the finer details and maintaining the overall structure of the objects. 
More qualitative results and comparisons are included in Figure \ref{fig:our_mesh_synthetic} and  \ref{fig: ablation}. 
\paragraph{Speed comparison}  
We compared the training and inference speed of our method with Neuralangelo~\cite{li2023neuralangelo} and 2DGS~\cite{Huang2DGS2024}, as shown in Table~\ref{tab: speed}. 
When training on the DTU dataset, neural surface reconstruction methods like Neuralangelo~\cite{li2023neuralangelo} require over 30 hours to produce the final results, while Gaussian-splatting-based methods like 2DGS~\cite{Huang2DGS2024} take only 18.8 minutes. 
Our method, 3DGSR, takes {30 minutes}-slightly longer than 2DGS~\cite{Huang2DGS2024}, but significantly faster than Neuralangelo~\cite{li2023neuralangelo}.
This discrepancy arises because 3DGSR incorporates geometric constraints that effectively optimize the implicit representation, though querying the rays during training incurs additional time costs. 
However, when rendering an 800x600 image on the DTU dataset, both 2DGS~\cite{Huang2DGS2024} and our method achieve a frame rate of 175 fps.
This is because surface regularization is applied solely during the training phase, allowing for the simultaneous learning of high-quality surfaces and Gaussians. 
Once the optimization is complete, the surface can be extracted offline and reused for future applications.

\begin{table}[t]
    \centering 
    \begin{tabular}{lcc}
        \toprule
        Method & Training Speed & Inference \\ 
        \hline
        Neuralangelo~\cite{li2023neuralangelo} & > 30 h & < 1 FPS \\
        2DGS~\cite{Huang2DGS2024}  & 18.8 min & 175 FPS  \\
        Ours  & 30 min   & 175 FPS  \\
        \bottomrule
    \end{tabular}
    \caption{Training and inference speed comparison on DTU dataset.}
    \label{tab: speed}
\end{table}
\begin{table}[t]
    \centering 
    \begin{tabular}{lccccc}
        \bottomrule[0.9pt]
        Method & C-$\mathcal{L}_1$ & PSNR & LPIPS & SSIM \\ 
        \hline
       \emph{static $\beta$ (Tight coupling})             & 1.95  & 33.02 & 0.0385  & 0.966 \\
         \emph{Poisson (Tight coupling})                    & 2.03   & -     & -       & - \\
        \emph{Ours (Tight coupling})          & {1.50}  &33.23& {0.0340} & {0.968}\\
        \emph{Ours (Loose coupling)}           & \textbf{1.37} &\textbf{33.86}& \textbf{0.0330} & \textbf{0.971}\\ \hline
         \textit{w/o depth,normal}                     & 3.63    & 32.83 & 0.0380  & 0.963 \\
          \textit{w/o  normal}                     & 1.74  & 33.18 & 0.0383  & 0.967 \\
        \toprule[0.9pt]
    \end{tabular}
    \caption{Ablation study of the surface and GS alignment on the NeRF Synthetic dataset and the volumetric rendering depth (normal) loss, where we evaluate the effect of each proposed component. C-$\mathcal{L}_1$ is an abbreviation for the Chamfer-$\mathcal{L}_1$ distance.}
    \label{tab: ablation}
\end{table}

\subsection{Ablation Studies} 
We conduct ablation studies on the NeRF Synthetic dataset in Table~\ref{tab: ablation} to evaluate the effectiveness of each module in our method. Four different configurations are investigated to train our model. 
We also show the qualitative results in Fig.~\ref{fig: ablation}.
\paragraph{Volumetric Constraints.} 
The volumetric constraints have a minimal impact on novel view synthesis (PSNR: 32.83-33.86) but significantly influence the reconstruction results (Chamfer-$\mathcal{L}_1$: 3.63-1.37). This is because the volumetric rendering for depth and normal supervision can provide dense supervisory signals for learning the continuous SDF. Without these constraints, numerous noisy structures would be generated in areas not occupied by Gaussians, negatively affecting the overall reconstruction quality.
As shown in Fig.~\ref{fig: ablation}, if we directly use the tight coupling strategy without any geometry constraint, we can only get the coarse result.
\paragraph{Reconstruction Strategy.} {Compared with the marching cubes on the SDF field for surface extraction, we also employ Poisson reconstruction to extract surface from Gaussians with normals derived from the implicit SDF, which yields a much lower performance.}
This demonstrates the advantage of using learned implicit SDF to represent the surface that is continuous, more tolerant to noisy supervisions, and allows for easy extraction of high-fidelity surfaces.  
Notably, the surface is modeled as a hash grid and a single-layer MLP, which is lightweight and does not introduce many burdens for computation. 
\paragraph{Coupling Strategy.}  
We experiment with the tight coupling and loose coupling strategies proposed by our approach. For the tight coupling strategy, the learning of the parameter $\beta$ impacts our performance. In contrast, the loose coupling strategy leads to improved results in both reconstruction and novel view synthesis, demonstrating its effectiveness in balancing the different aspects of the modeling process.
\begin{figure}
	\centering
	\includegraphics[width=0.95\linewidth]{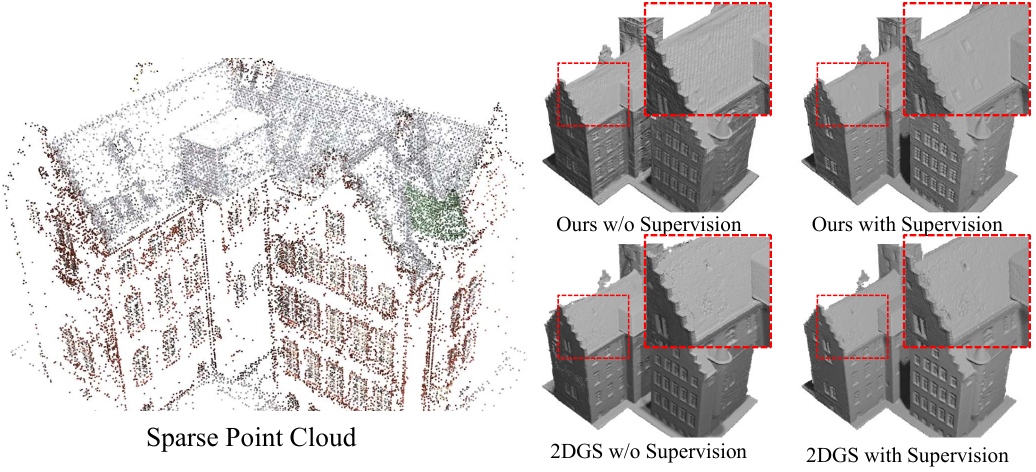}
	\caption{Sparse point cloud and the influence of projected depth supervision $D_{sfm}$ in 2DGS and 3DGSR.}
\label{fig: Depth}
\end{figure}
\paragraph{Projected Sparse Depth $D_{sfm}$.}

\begin{table}[t]
    \centering 
    \begin{tabular}{lcc}
        \toprule
        Method & $D_{sfm}$ & Chamfer-$\mathcal{L}_1$ \\ 
        \hline
        2DGS~\cite{Huang2DGS2024} & {\textit{Yes}}   & 0.85 \\
        2DGS~\cite{Huang2DGS2024} & {\textit{No}} & 0.80 \\
        Ours & {\textit{Yes}}   & {0.70} \\
        Ours & {\textit{No}} & {0.81} \\
        \bottomrule
    \end{tabular}
    \caption{Ablation study of sparse depth supervision. {\textit{Yes}} indicates the use of \( D_{sfm} \) as a supervision signal, while {\textit{No}} indicates that \( D_{sfm} \) is not used for supervision.}
    \label{tab: depth supervision}
\end{table}

To ensure a fair comparison, we conducted experiments to analyze the influence of the supervision by projected sparse depth $D_{sfm}$ on both 3DGSR and 2DGS~\cite{Huang2DGS2024}, presenting the quantitative results in Table~\ref{tab: depth supervision} and qualitative results in Fig.~\ref{fig: Depth}.
As shown in Table~\ref{tab: depth supervision}, we observed that adding sparse depth cues negatively affects the performance of 2DGS~\cite{Huang2DGS2024}, causing a worsening in Chamfer-$\mathcal{L}_1$ from 0.80 to 0.85. This performance drop may stem from the fact that 2DGS~\cite{Huang2DGS2024} relies on depth fusion for reconstruction, as opposed to our approach, which jointly optimizes and aligns the surface with the 3D Gaussians.
In contrast, the addition of sparse depth supervision proves beneficial in our method, resulting in improved quantitative results (Chamfer-$\mathcal{L}_1$ from 0.81 to 0.70). These cues assist SDF optimization in regions where image supervision is sparse and camera poses are inaccurate, though they are less impactful in areas where sufficient observations are available. 
When trained using only image signals, our method can be comparable with 2DGS~\cite{Huang2DGS2024}.
Fig.~\ref{fig: Depth} further demonstrates that 2DGS~\cite{Huang2DGS2024} shows only minor improvements with sparse depth supervision, whereas our method achieves more detailed reconstructions even in the absence of such supervision.
In conclusion, these depth cues are valuable for improving performance in cases of inaccurate camera poses and ill-posed sparse view data, especially in real-world scenes. 
Our approach can effectively utilize these free cues to enhance reconstruction quality.

%% file: main/6.conclusion.tex
\section{Conclusion}

In this paper, we introduce 3DGSR, an efficient method for high-quality surface reconstruction using 3DGS. Our approach is based on two key components: 1) the integration of neural implicit SDF and its alignment with 3DGS, and 2) the utilization of volumetric rendering and the SDF-and-Gaussian geometry consistency regularization for SDF optimization. The first component allows for the joint optimization of SDF and 3D Gaussians, with the optimization of Gaussians providing supervision signals for learning the SDF. The second component supplies additional supervision signals for refining the SDF in areas not occupied by Gaussians, using a consistency loss that aligns depth (normal) from SDF with that derived from Gaussians. Our extensive experiments showcase the effectiveness of 3DGSR in reconstructing high-quality surfaces outperforming those obtained from state-of-the-art reconstruction pipelines, without compromising the rendering capabilities and efficiency of 3D Gaussians. We hope our approach could inspire more future work. 
Despite its promising results, our approach still has limited capability to handle unbounded scenes, due to the use of a hash grid and implicit SDF, and also has worse results on real datasets with noisy pose estimation results. 
Furthermore, it may encounter difficulties in reconstructing transparent objects. These limitations highlight potential areas for further improvement and development in our method.

\section{Acknowledgement}

This work has been supported by the Hong Kong Research Grant Council - Early Career Scheme (Grant No. 27209621), General Research Fund Scheme (Grant No. 17202422, 17212923), Theme-based Research (Grant No. T45-701/22-R) and RGC Matching Fund Scheme (RMGS). Part of the described research work is conducted in the JC STEM Lab of Robotics for Soft Materials funded by The Hong Kong Jockey Club Charities Trust.

%% file: main/figure_only.tex
\appendix 
\begin{figure*}
	\centering
	\includegraphics[width=1.0\linewidth]{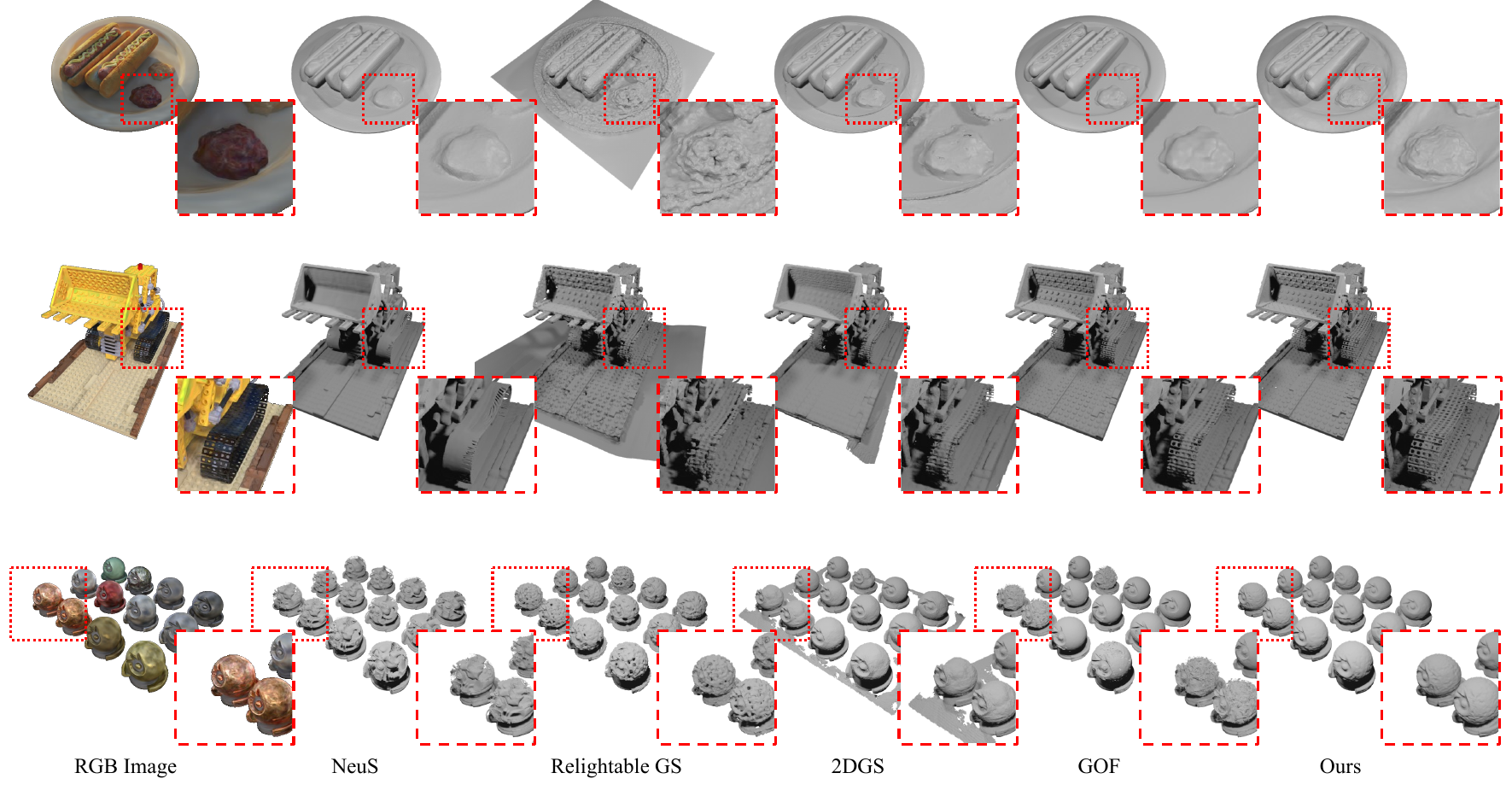}
	\caption{Qualitative comparison on NeRF-synthetic.
 }
\label{fig:our_mesh_comparison}
\end{figure*}

\begin{figure*}
	\centering
	\includegraphics[width=1.\linewidth]{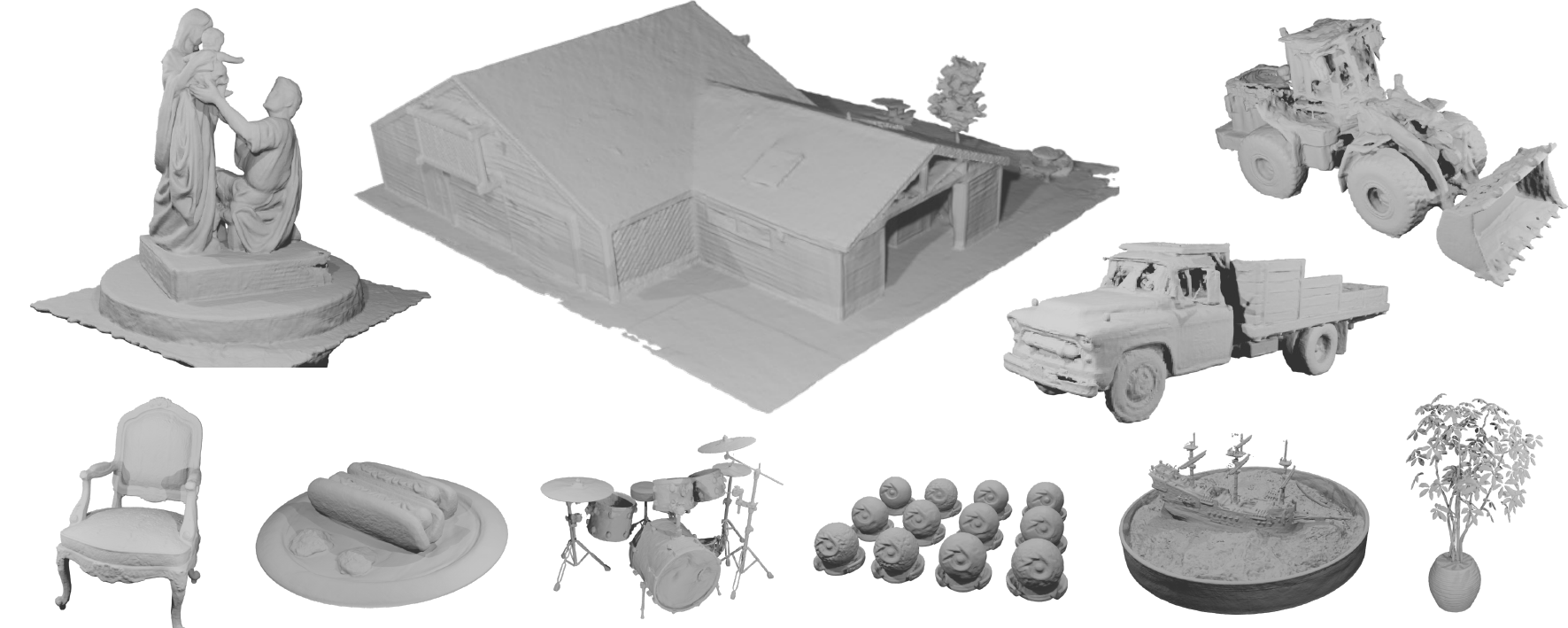}
	\caption{Qualitative results of our 3DGSR on the NeRF synthetic dataset~\cite{mildenhall2021nerf} and TNT dataset~\cite{knapitsch2017tanks}.}
\label{fig:our_mesh_synthetic}
\end{figure*}

\begin{figure*}
	\centering
	\includegraphics[width=0.9\linewidth]{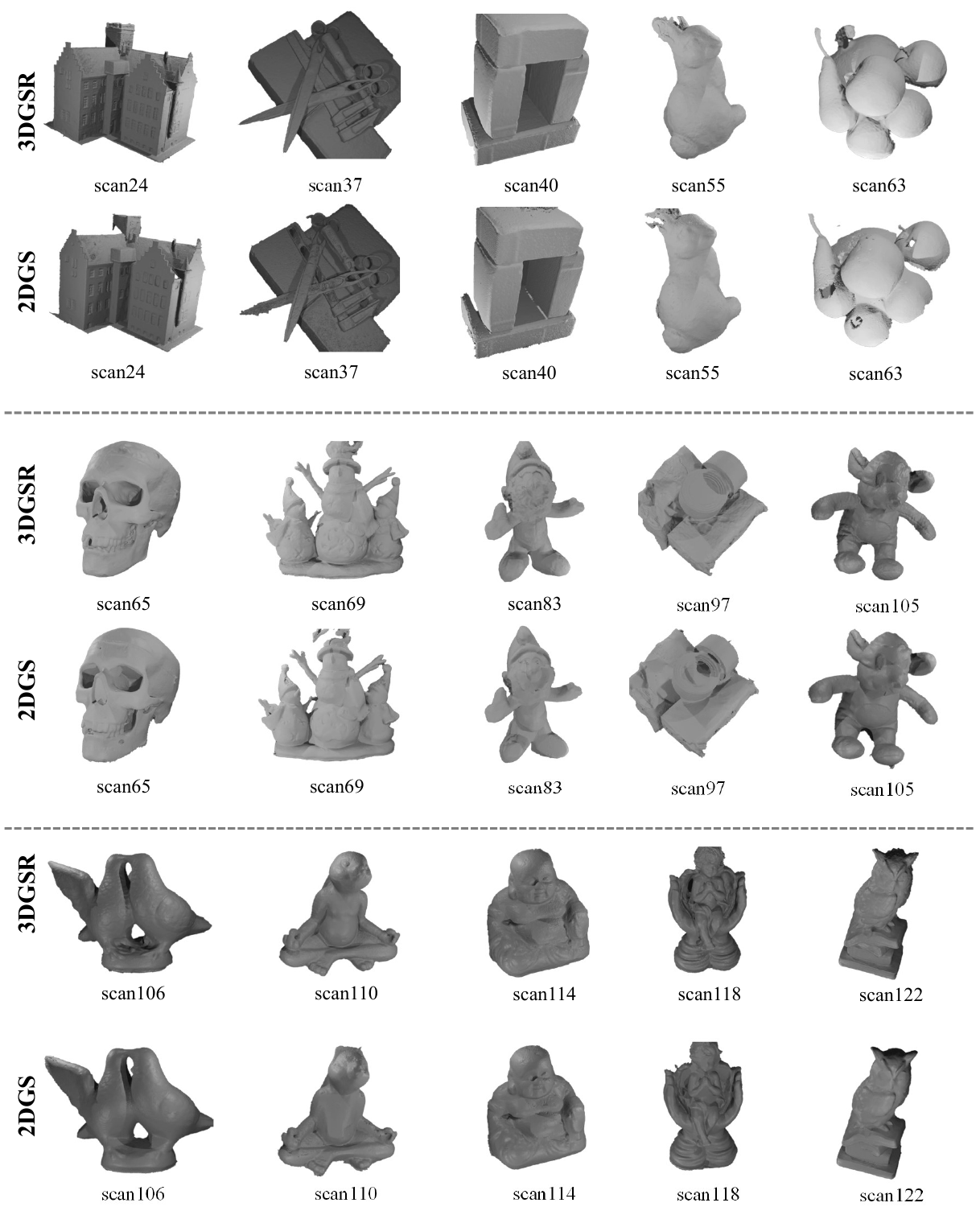}
	\caption{Comparison of surface reconstruction using our 3DGSR and 2DGS~\cite{Huang2DGS2024}.}
\label{fig: ablation}
\end{figure*}